%% file: main.tex
\documentclass{article} % For LaTeX2e
\usepackage[dvipsnames]{xcolor}

\usepackage{iclr2021_conference,times}

\input{math_commands.tex}

\usepackage[utf8]{inputenc} % allow utf-8 input
\usepackage[T1]{fontenc}    % use 8-bit T1 fonts
\usepackage{hyperref}       % hyperlinks
\usepackage{url}            % simple URL typesetting
\usepackage{booktabs}       % professional-quality tables
\usepackage{amsfonts}       % blackboard math symbols
\usepackage{nicefrac}       % compact symbols for 1/2, etc.
\usepackage{microtype}      % microtypography
\usepackage{graphicx}
\usepackage{amsmath,amssymb} % define this before the line numbering.
\usepackage{color}
\usepackage{epsfig}
\usepackage{tabularx}
\usepackage{booktabs}
\usepackage{multirow}
\usepackage{array}
\usepackage{xspace}
\usepackage{balance}
\usepackage{enumitem}
\usepackage{mathtools}
\usepackage{rotating}
\usepackage{tabulary}
\usepackage{subcaption}
\usepackage[export]{adjustbox}
\usepackage{ctable}
\usepackage{diagbox}
\usepackage{subcaption}
\usepackage{makecell}
\usepackage[linesnumbered,ruled]{algorithm2e}
\usepackage{bbm}
\usepackage[title]{appendix}

\newcommand{\ie}{\textit{i}.\textit{e}.}
\newcommand{\eg}{\textit{e}.\textit{g}.}

\graphicspath{{./figs/}}

\title{Prototypical Contrastive Learning of\\Unsupervised Representations} % Replace with your title

\author{Junnan Li, Pan Zhou, Caiming Xiong, Steven C.H. Hoi\\
	Salesforce Research\\
	\texttt{\{junnan.li,pzhou,cxiong,shoi\}@salesforce.com}
}

\iclrfinalcopy

\begin{document}
	\maketitle

   \input{sec_abstract}
   \input{sec_introduction}
   \input{sec_literature}

   \input{sec_method}

   \input{sec_experiment}
   \input{sec_conclusion}

\bibliographystyle{iclr2021_conference}
\bibliography{bib}

\input{sec_appendix}

\end{document}

%% file: math_commands.tex
\usepackage{amsmath,amsfonts,bm}

\def\eqref#1{equation~\ref{#1}}
\def\1{\bm{1}}

\DeclareMathAlphabet{\mathsfit}{\encodingdefault}{\sfdefault}{m}{sl}
\SetMathAlphabet{\mathsfit}{bold}{\encodingdefault}{\sfdefault}{bx}{n}

\DeclareMathOperator*{\argmax}{arg\,max}
\DeclareMathOperator*{\argmin}{arg\,min}

%% file: sec_abstract.tex
\vspace{-1ex}
\begin{abstract}
This paper presents Prototypical Contrastive Learning (PCL),
an unsupervised representation learning method that bridges contrastive learning with clustering.
PCL not only learns low-level features for the task of instance discrimination,
but more importantly,
it encodes semantic structures discovered by clustering into the learned embedding space.
%and prevents the network from solely relying on low-level cues for solving unsupervised learning tasks. 
Specifically,
we introduce prototypes as latent variables to help find the maximum-likelihood estimation of the network parameters in an Expectation-Maximization framework.
We iteratively perform E-step as finding the distribution of prototypes via clustering and M-step as optimizing the network via contrastive learning.
We propose ProtoNCE loss,
a generalized version of the InfoNCE loss for contrastive learning,
which encourages representations to be closer to their assigned prototypes.
PCL outperforms state-of-the-art instance-wise contrastive learning methods on multiple benchmarks with substantial improvement in low-resource transfer learning.
Code and pretrained models are available at \textcolor{magenta}{\url{https://github.com/salesforce/PCL}}.
\end{abstract}

%% file: sec_introduction.tex
\section{Introduction}
\label{sec:introduction}
\vspace{-1ex}
Unsupervised visual representation learning aims to learn image representations from pixels themselves without relying on semantic annotations, and 
recent advances are largely driven by instance discrimination tasks~\citep{Instance,end2end,moco,PIRL,DMI,CPC,CMC}.
These methods 
%aim to learn a deep embedding space which preserves the visual similarity of image instances.
usually consist of two key components: image transformation and contrastive loss.
Image transformation aims to generate multiple embeddings that represent the same image, by data augmentation~\citep{end2end,AMDIM,simple}, patch perturbation~\citep{PIRL}, or using momentum features~\citep{moco}.
The contrastive loss, in the form of a noise contrastive estimator~\citep{NCE},
aims to bring closer samples from the same instance and separate samples from different instances.
Essentially,
instance-wise contrastive learning leads to an embedding space where all instances are well-separated,
and each instance is locally smooth (\ie~input with perturbations have similar representations).

Despite their improved performance, instance discrimination methods share a common weakness: the representation is not encouraged to encode the semantic structure of data. This problem arises because instance-wise contrastive learning treats two samples as a negative pair as long as they are from different instances,
regardless of their semantic similarity. 
%between instances.
This is magnified by the fact that thousands of negative samples are generated to form the contrastive loss,
leading to many negative pairs that share similar semantics but are undesirably pushed apart in the embedding space.

\input{table/fig_illustration}

In this paper,
we propose \textit{prototypical contrastive learning} (PCL),
a new framework for unsupervised representation learning that implicitly encodes the semantic structure of data into the embedding space.
Figure~\ref{fig:illustration} shows an illustration of PCL.
A prototype is defined as ``a representative embedding for a group of semantically similar instances''.
We assign several prototypes of different granularity to each instance,
and construct a contrastive loss which enforces the embedding of a sample to be more similar to its corresponding prototypes compared to other prototypes.
In practice, we can find prototypes by performing clustering on the embeddings.

We formulate prototypical contrastive learning as an Expectation-Maximization (EM) algorithm,
where the goal is to find the parameters of a Deep Neural Network (DNN) that best describes the data distribution,
by iteratively approximating and maximizing the log-likelihood function. Specifically, 
we introduce prototypes as additional latent variables,
and estimate their probability in the E-step by performing $k$-means clustering.
In the M-step, we update the network parameters by minimizing our proposed contrastive loss, namely \textit{ProtoNCE}. 
We show that minimizing ProtoNCE is equivalent to maximizing the estimated log-likelihood,
under the assumption that the data distribution around each prototype is isotropic Gaussian.
Under the EM framework,
the widely used instance discrimination task can be explained as a special case of prototypical contrastive learning,
where the prototype for each instance is its augmented feature, %(\ie~number of prototypes = number of instances),
and the Gaussian distribution around each prototype has the same fixed variance.  The contributions of this paper can be summarized as follows:
\vspace{-\topsep}
\begin{itemize}[leftmargin=*]
	\setlength\itemsep{0pt}
	\item 
	We propose prototypical contrastive learning, a novel framework for unsupervised representation learning that bridges contrastive learning and clustering.
	The learned representation is encouraged to capture the hierarchical semantic structure of the dataset.
	\item
	We give a theoretical framework that formulates PCL as an Expectation-Maximization (EM) based algorithm.
	The iterative steps of clustering and representation learning can be interpreted as approximating and maximizing the log-likelihood function.
	The previous methods based on instance discrimination form a special case in the proposed EM framework.
	%Instance-wise contrastive learning is explained as a special case of prototypical contrastive learning.
	\item 
	We propose ProtoNCE, 
	a new contrastive loss which improves the widely used InfoNCE by dynamically estimating the concentration for the feature distribution around each prototype. 
ProtoNCE also includes an InfoNCE term in which the instance embeddings can be interpreted as instance-based prototypes.
	We provide explanations for PCL from an information theory perspective,
	by showing that the learned prototypes contain more information about the image classes.
	\item 
    PCL outperforms instance-wise contrastive learning on multiple benchmarks with substantial improvements in low-resource transfer learning.    
    PCL also leads to better clustering results.
    %Prototypical contrastive learning is compatible to techniques proposed to improve
\end{itemize}

%From an information theory perspective,
%prototypical contrastive learning also shows advantage compared to instance-wise contrastive learning.

%In practice, we perform k-means clustering on the embeddings, and use the cluster centroids as prototypes.
%The granularity of the prototypes (\ie~the size of the clusters) can be controlled by the total number of clusters.
%In practice, we find prototypes by performing standard clustering (\eg~$k$-means) on the embeddings,
%and use the cluster centroids as prototypes.
%The granularity of a prototype is inversely proportional to the total number of prototypes.

%% file: table/fig_illustration.tex
\begin{figure*}[!t]
 \centering
 \vspace{-0.15in}
   \includegraphics[width=0.9\linewidth]{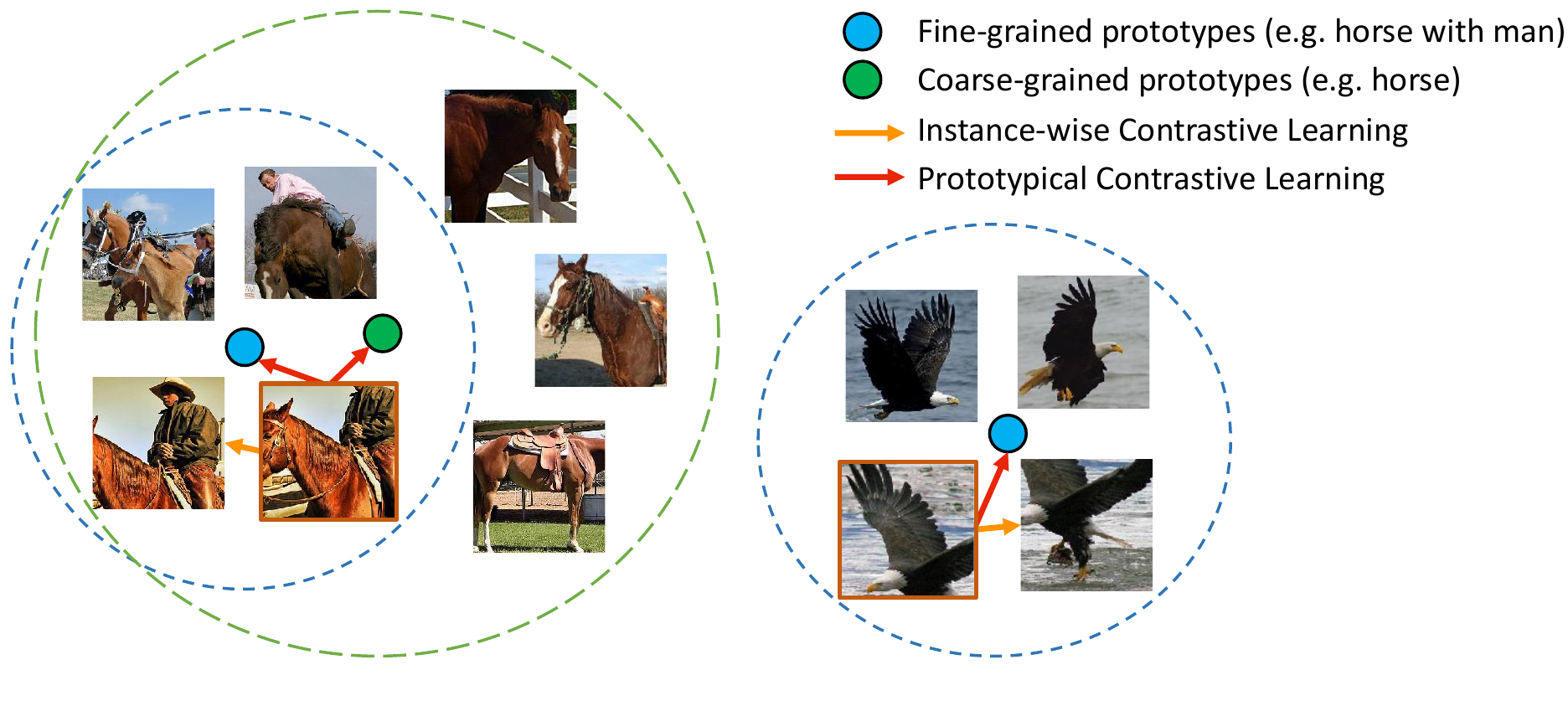}
   \vspace{-0.1in}
   \caption
  	{
  	\small
		Illustration of Prototypical Contrastive Learning. Each instance is assigned to multiple prototypes with different granularity.
		PCL learns an embedding space which encodes the semantic structure of data.
	} 
	\vspace{-0.15in}
  \label{fig:illustration}
 \end{figure*}

%% file: sec_literature.tex
\vspace{-2ex}
\section{Related work}
\label{sec:literature}
\vspace{-1ex}

Our work is closely related to two main branches of studies in unsupervised/self-supervised learning: instance-wise contrastive learning and deep unsupervised clustering.

%In this section, we review related work and highlight our differences.

%\subsection{Instance-wise Contrastive Learning}
\vspace{-0.5ex}
{\bf Instance-wise contrastive learning}~\citep{Instance,end2end,moco,PIRL,local,DMI,CPC,CMC,simple} 
aims to learn an embedding space where samples (\eg~crops) from the same instance (\eg~an image) are pulled closer and samples from different instances are pushed apart.
%Such unsupervised learned embedding would preserve the visual similarity of images.
To construct the contrastive loss, % for a mini-batch of samples,
positive instance features and negative instance features are generated for each sample. 
Different contrastive learning methods vary in their strategy to generate instance features.
The \textit{memory bank} approach~\citep{Instance} stores the features of all samples calculated in the previous step.
The \textit{end-to-end} approach~\citep{end2end,CMC,simple} generates instance features using all samples within the current mini-batch.
%and apply the same encoder to both the original samples and their augmented version.
The \textit{momentum encoder} approach~\citep{moco} encodes samples on-the-fly by a momentum-updated encoder,
and maintains a queue of instance features.

Despite their improved performance, the existing
methods based on instance-wise contrastive learning have the following two major limitations, which can be addressed by the proposed PCL framework. 
\vspace{-\topsep}
\begin{itemize}[leftmargin=*]
	\setlength\itemsep{0pt}
	\item The task of instance discrimination could be solved by exploiting low-level image differences, thus the learned embeddings do not necessarily capture high-level semantics. This is supported by the fact that the accuracy of instance classification often rapidly rises to a high level (>90\% within 10 epochs) and further training gives limited informative signals.
	A recent study also shows that better performance of instance discrimination could worsen the performance on downstream tasks~\citep{MI_limitation}.
	\item A sufficiently large number of negative instances need to be sampled, which inevitably yields negative pairs that share similar semantic meaning and should be closer in the embedding space. However, they are undesirably pushed apart by the contrastive loss.
	Such problem is defined as class collision in~\citep{Saunshi_ICML_2019} and is shown to hurt representation learning.
	Essentially, instance discrimination learns an embedding space that only preserves the local smoothness around each instance but largely ignores the global semantic structure of the dataset.	
\end{itemize}

%\subsection{Deep Unsupervised Clustering} 

{\bf Deep unsupervised clustering.} Clustering based methods have been proposed for deep unsupervised learning.
\cite{Xie_ICML_2016,Yang_CVPR_2016,Liao_NIPS_2016,Yang_ICML_2017,DAC,Invariant_cluster,SCAN} jointly learn image embeddings and cluster assignments,
but they have not shown the ability to learn transferable representations from a large scale of images.
Closer to our work,
DeepCluster~\citep{deepcluster} performs iterative clustering and unsupervised representation learning,
which is further improved by~\cite{online} with online clustering.
However, our method is conceptually different from DeepCluster.
In DeepCluster, the cluster assignments are considered as pseudo-labels and a classification objective is optimized, which results in two weaknesses:
(1) the high-dimensional features from the penultimate layer of a ConvNet are not optimal for clustering and need to be PCA-reduced;
(2) an additional linear classification layer is frequently re-initialized which interferes with representation learning.
In our method,
representation learning happens directly in a low-dimensional embedding space,
by optimizing a contrastive loss on the prototypes (cluster centroids).
Concurrent to our work,
SwAV~\citep{swav} also brings together a clustering objective with contrastive learning.
%However, their clustering is performed on a subset of the data.
% This frees our method from the computationally expensive linear layer,
% and enables a much larger number of clusters where each instance is assigned to multiple clusters of different granularity.
%Furthermore,
%we perform clustering using features from the momentum encoder, which produces more consistent clusters.

%\subsection{Self-supervised Pretext Tasks} 

{\bf Self-supervised pretext tasks.} Another line of self-supervised learning methods focus on training DNNs to solve pretext tasks,
%that lead to good image representations being learned.
which usually involve hiding certain information about the input and training the network to recover those missing information.
Examples include image inpainting~\citep{Pathak_CVPR_2016}, colorization~\citep{Zhang_ECCV_2016,Zhang_CVPR_2017},
prediction of patch orderings~\citep{Doersch_ICCV_2015,Noroozi_ECCV_2016} and image transformations~\citep{exampler,Gidaris_2018_ICLR,deepcluster_rotation,AET}.  
% However, most of these pretext tasks exploit specific structures of visual data,
% making them harder to generalize to other domains.
Compared to heuristic pretext task designs, the proposed PCL is a more general learning framework with better theoretical justification.
% Furthermore,
% PCL can incorporate the pretext tasks (\eg~ Jigsaw~\citep{Noroozi_ECCV_2016} or Rotation~\citep{Gidaris_2018_ICLR}) as a form of image transformation,
% which could potentially lead to improved performance.

%% file: sec_method.tex
\section{Prototypical Contrastive Learning}
\label{sec:method}

\subsection{Preliminaries}
Given a training set $X=\{x_1, x_2,...,x_n\}$ of $n$ images,
unsupervised visual representation learning aims to learn an embedding function $f_{\theta}$ (realized via a DNN) that maps $X$ to $V=\{v_1, v_2,...,v_n\}$ with $v_i=f_{\theta}(x_i)$,
such that $v_i$ best describes $x_i$.
Instance-wise contrastive learning achieves this objective by optimizing a contrastive loss function, such as InfoNCE~\citep{CPC,moco}, defined as:
\begin{equation}
\label{eqn:infoNCE}
	\mathcal{L}_\mathrm{InfoNCE} = \sum_{i=1}^n  -\log \frac{\exp(v_i\cdot v_i'/\tau)}{\sum_{j=0}^r \exp(v_i\cdot v_j'/\tau) } ,
\end{equation}
where $v_i$ and $v_i'$ are positive embeddings for instance $i$,
and $v_j'$ includes one positive embedding and $r$ negative embeddings for other instances, and $\tau$ is a temperature hyper-parameter. 
In MoCo~\citep{moco}, these embeddings are obtained by feeding $x_i$ to a momentum encoder parametrized by $\theta'$, $v_i'=f_{\theta'}(x_i)$,
where $\theta'$ is a moving average of $\theta$.

In prototypical contrastive learning, 
we use prototypes $c$ instead of $v'$,
and replace the fixed temperature $\tau$ with a per-prototype concentration estimation $\phi$.
An overview of our training framework is shown in Figure~\ref{fig:framework},
where clustering and representation learning are performed iteratively at each epoch.
Next,
we will delineate the theoretical framework of PCL based on EM.
A pseudo-code of our algorithm is given in appendix~\ref{sec:pseudo_code}.

%In prototypical contrastive learning,
%we replace $v'$ with prototypes $c$,
%and change the fixed temperature hyper-parameter $\tau$ into a concentration measure $\phi$ that is estimated from the data.
%Next we introduce the framework of our method and delineate the details.
\input{table/fig_framework}
\subsection{PCL as Expectation-Maximization}
\label{sec:em}

Our objective is to find the network parameters $\theta$ that maximizes the log-likelihood function of the observed $n$ samples:
\vspace{-0.4em}
\begin{equation}
	\theta^* = \argmax_\theta \sum_{i=1}^n \log p(x_i;\theta)
\end{equation}
We assume that the observed data $\{x_i\}_{i=1}^n$ are related to latent variable $C=\{c_i\}_{i=1}^k$ which denotes the prototypes of the data.  In this way, 
we can re-write the log-likelihood function as:
\vspace{-0.4em}
\begin{equation}
\label{eqn:mle}
\theta^* = \argmax_\theta \sum_{i=1}^n \log p(x_i;\theta) = \argmax_\theta \sum_{i=1}^n \log \sum_{c_i \in C}  p(x_i,c_i;\theta) 
\end{equation}
It is hard to optimize this function directly,
so we use a surrogate function to lower-bound it:
\vspace{-0.4em}
\begin{equation}
\label{eqn:jensen}
\begin{split}
\sum_{i=1}^n \log \sum_{c_i \in C}  p(x_i,c_i;\theta) &= 
\sum_{i=1}^n \log \sum_{c_i \in C} Q(c_i) \frac{p(x_i,c_i;\theta)}{Q(c_i)}  \geq \sum_{i=1}^n \sum_{c_i \in C} Q(c_i)  \log \frac{p(x_i,c_i;\theta)}{Q(c_i)},
\end{split}
\end{equation}
where $Q(c_i)$ denotes some distribution over $c$'s ($\sum_{c_i \in C} Q(c_i)=1$),
and the last step of derivation uses Jensen's inequality.
To make the inequality hold with equality, 
we require $\small{\frac{p(x_i,c_i;\theta)}{Q(c_i)}}$ to be a constant.
Therefore, we have:
\vspace{-0.4em}
\begin{equation}
Q(c_i) = \frac{p(x_i,c_i;\theta)}{\sum_{c_i \in C} p(x_i,c_i;\theta)}
          = \frac{p(x_i,c_i;\theta)}{p(x_i;\theta)}
          = p(c_i ; x_i,\theta)
\end{equation}
By ignoring the constant $-\sum_{i=1}^n \sum_{c_i \in C} Q(c_i) \log Q(c_i)$ in eqn.(\ref{eqn:jensen}),
we should maximize:
\begin{equation}
\label{eqn:bound}
\sum_{i=1}^n  \sum_{c_i \in C} Q(c_i) \log p(x_i,c_i;\theta)
\end{equation}
\noindent \textbf{E-step.} 
In this step, we aim to estimate $p(c_i ; x_i,\theta)$.
To this end,
we perform $k$-means  on the features $v_i' = f_{\theta'}(x_i)$ given by the momentum encoder to obtain $k$ clusters. We define prototype $c_i$ as the centroid for the $i$-th cluster.
Then, we %estimate $p(c_i ; x_i,\theta)$ as
compute $p(c_i ; x_i,\theta)= \mathbbm{1} (x_i \in c_i)$,
where $\mathbbm{1} (x_i \in c_i)=1$ if $x_i$ belongs to the cluster represented by $c_i$;
otherwise $\mathbbm{1} (x_i \in c_i)=0$.
Similar to MoCo,
we found features from the momentum encoder yield more consistent clusters.

% by solving the lower-level optimization in Equation~\ref{eqn:bilivel}.
%$C$ is defined as cluster centroids and $S$ is defined as cluster assignments.
%We find that using features from the momentum encoder generates more consistent clusters.
%The $k$-means clustering can be interpreted as a special case of a truncated EM applied to Gaussian Mixture Models (GMM) with isotropic Gaussians~\citep{kmeans_em}.
%Hence our upper-level E-step contains another inner-loop of EM.

%In order to obtain prototypes of different granularities,
%we perform multiple rounds of $k$-means clustering with a set of $K=\{K_1,..,K_Q\}$,
%which leads to multiple sets of prototypes and assignments, $\{(C_q,S_q)\}_{q=1}^Q$.
%We assign each of them with an equal probability, $p(C_q,S_q)=1/Q$.

\noindent \textbf{M-step.} 
Based on the E-step, we are ready to maximize the lower-bound in eqn.(\ref{eqn:bound}).
\begin{equation}
\label{eqn:mstep}
\begin{split}
\sum_{i=1}^n  \sum_{c_i \in C} Q(c_i) \log p(x_i,c_i;\theta) &
= \sum_{i=1}^n  \sum_{c_i \in C} p(c_i ; x_i,\theta) \log p(x_i,c_i;\theta)\\
 &= \sum_{i=1}^n  \sum_{c_i \in C} \mathbbm{1} (x_i \in c_i) \log p(x_i,c_i;\theta)
%& = \sum_{i=1}^n \log p(x_i,c_s;\theta) \\
\end{split}
\end{equation}
Under the assumption of a uniform prior over cluster centroids, we have:
\vspace{-0.4em}
\begin{equation}
\label{eqn:likelihood}
p(x_i,c_i;\theta) = p(x_i;c_i,\theta) p(c_i;\theta) = \frac{1}{k}\cdot p(x_i;c_i,\theta),
\end{equation}
where we set the prior probability $p(c_i;\theta)$ for each $c_i$ as $1/k$ since we are not provided any samples.

 We assume that the distribution around each prototype is an isotropic Gaussian, which leads to:
\begin{equation}
\label{eqn:gaussian}
p(x_i;c_i,\theta) = {\exp\left(\frac{-(v_i-c_s)^2}{2\sigma_s^2}\right)}\Big/{\sum_{j=1}^{k} \exp\bigg(\frac{-(v_i-c_j)^2}{2\sigma_j^2}\bigg)},
%p(x_i;c_i,\theta) = \frac{\exp(\frac{-(v_i-c_s)^2}{2\sigma_s^2})}{\sum_{j=1}^{k} \exp(\frac{-(v_i-c_j)^2}{2\sigma_j^2})},
\end{equation}
where $v_i=f_\theta(x_i)$ and $x_i \in c_s$.
%\begin{equation}
%\log p(v_i|C_q,S_q) = \log \frac{\exp(\frac{-(v_i-c_q^a)^2}{2\sigma_a^2})}{\sum_{b=1}^{K_q} \exp(\frac{-(v_i-c_q^b)^2}{2\sigma_b^2})},
%\end{equation}
%where $a=S_q^i$ is the assignment of $x_i$ to the $a$-th prototype in $C_q$.
If we apply $\ell_2$-normalization to both $v$ and $c$,
then $(v-c)^2=2-2 v\cdot c$.
Combining this with eqn.(\ref{eqn:mle},~\ref{eqn:jensen},~\ref{eqn:bound},~\ref{eqn:mstep},~\ref{eqn:likelihood},~\ref{eqn:gaussian}),
we can write maximum log-likelihood estimation as
\begin{equation}
\label{eqn:proto}
\theta^* = \argmin_\theta \sum_{i=1}^n  -\log \frac{\exp(v_i\cdot c_s/\phi_s)}{\sum_{j=1}^{k} \exp(v_i\cdot c_j/\phi_j)}, 
\end{equation}
where $\phi\propto\sigma^2$ denotes the concentration level of the feature distribution around a prototype and will be introduced later.
Note that eqn.(\ref{eqn:proto}) has a similar form as the InfoNCE loss in eqn.(\ref{eqn:infoNCE}).
Therefore, 
InfoNCE can be interpreted as a special case of the maximum log-likelihood estimation,
where the prototype for a feature $v_i$ is the augmented feature $v'_i$ from the same instance (\ie~$c=v'$),
and the concentration of the feature distribution around each instance is fixed (\ie ~$\phi=\tau$).

In practice,
we take the same approach as NCE and sample $r$ negative prototypes to calculate the normalization term.
We also cluster the samples $M$ times with different number of clusters $K=\{k_m\}_{m=1}^M$,
which enjoys a more robust probability estimation of prototypes that encode the hierarchical structure.
Furthermore,
we add the InfoNCE loss to retain the property of local smoothness and help bootstrap clustering.
Our overall objective,  namely \textbf{ProtoNCE}, is defined as 
\begin{equation}
\label{eqn:proto_NCE}
\mathcal{L}_\mathrm{ProtoNCE} = \sum_{i=1}^n  -\bigg(\log \frac{\exp(v_i\cdot v_i'/\tau)}{\sum_{j=0}^r \exp(v_i\cdot v_j'/\tau) }+\frac{1}{M}\sum_{m=1}^M \log \frac{\exp(v_i\cdot c_s^m/\phi_s^m)}{\sum_{j=0}^{r} \exp(v_i\cdot c_j^m/\phi_j^m)} \bigg).
\end{equation}

\vspace{-0.1in}
\subsection{Concentration estimation}

The distribution of embeddings around each prototype has different level of concentration. %as illustrated in Figure~\ref{fig:illustration}.
We use $\phi$ to denote the concentration estimation, where a smaller $\phi$ indicates larger concentration.
Here we calculate $\phi$ using the momentum features $\{v'_z\}_{z=1}^Z$ that are within the same cluster as a prototype $c$. 
The desired $\phi$ should be small (high concentration) if (1) the average distance between $v'_z$ and $c$ is small, and (2) the cluster contains more feature points (\ie~$Z$ is large).
Therefore, we define $\phi$ as:
\begin{equation}
\label{eqn:mu}
\phi = \frac{\sum_{z=1}^Z \lVert v'_z-c\rVert_2}{Z \log(Z+\alpha)}, 
\end{equation}
where $\alpha$ is a smooth parameter to ensure that small clusters do not have an overly-large $\phi$.
We normalize $\phi$ for each set of prototypes $C^m$ such that they have a mean of $\tau$.

In the ProtoNCE loss (eqn.(\ref{eqn:proto_NCE})), 
$\phi_s^m$ acts as a scaling factor on the similarity between an embedding $v_i$ and its prototype $c_s^m$.
With the proposed $\phi$,
the similarity in a loose cluster (larger $\phi$) are down-scaled,
pulling embeddings closer to the prototype. 
On the contrary,
embeddings in a tight cluster (smaller $\phi$) have an up-scaled similarity, thus less encouraged to approach the prototype.
Therefore, learning with ProtoNCE yields more balanced clusters with similar concentration, as shown in Figure~\ref{fig:cluster_MI}(a).
It prevents a trivial solution where most embeddings collapse to a single cluster,
a problem that could only be heuristically addressed by data-resampling in DeepCluster~\citep{deepcluster}.
%Furthermore, since most datasets of natural images, such as ImageNet~\citep{ImageNet}

\input{table/fig_cluster_MI}

\vspace{-0.1in}
\subsection{Mutual information analysis}
It has been shown that minimizing InfoNCE is maximizing a lower bound on the mutual information (MI) between representations $V$ and $V'$~\citep{CPC}.
Similarly,
minimizing the proposed ProtoNCE can be considered as simultaneously maximizing the mutual information between $V$ and all the prototypes $\{V',C^1,...,C^M\}$.
This leads to better representation learning, for two reasons.

First, the encoder would learn the \textit{shared} information among prototypes,
and ignore the individual noise that exists in each prototype.
The shared information is more likely to capture higher-level semantic knowledge. Second, we show that \textit{compared to instance features, prototypes have a larger mutual information with the class labels}.
We estimate the mutual information between the instance features (or their assigned prototypes) and the ground-truth class labels for all images in ImageNet~\citep{ImageNet} training set,
following the method in~\citep{ross2014mutual}.
We compare the obtained MI of our method (ProtoNCE) and that of MoCo (InfoNCE).
As shown in Figure~\ref{fig:cluster_MI}(b),
compared to instance features,
the prototypes have a larger MI with the class labels due to the effect of clustering.
Furthermore,
compared to InfoNCE,
training on ProtoNCE can increase the MI of prototypes as training proceeds,
indicating that better representations are learned to form more semantically-meaningful clusters. 

\vspace{-0.05in}
\subsection{Prototypes as linear classifier}
\vspace{-0.05in}
Another interpretation of PCL can provide more insights into the nature of the learned prototypes.
The optimization in eqn.(\ref{eqn:proto}) is similar to optimizing the cluster-assignment probability $p(s;x_i,\theta)$ using the cross-entropy loss,
where the prototypes $c$ represent weights for a linear classifier.
With $k$-means clustering,
the linear classifier has a fixed set of weights as the mean vectors for the representations in each cluster, $c =\frac{1}{Z} \sum_{z=1}^Z v_z'$.
A similar idea has been used for few-shot learning~\citep{protonet},
where a non-parametric prototypical classifier performs better than a parametric linear classifier.

%Our code and models will be released.

\if 0 
We use SGD as our optimizer, with a weight decay of 0.0001, a momentum of 0.9, and a batch size of 256.
We train for 200 epochs, where we warm-up the network in the first 20 epochs by only using the InfoNCE loss.
The initial learning rate is 0.03, and is multiplied by 0.1 at 120 and 160 epochs.
In terms of the hyper-parameters,
we set $\tau=0.1$,
$\alpha=10$, and number of clusters $K=\{25000,50000,100000\}$.
We find over-clustering to be beneficial.
We use the GPU $k$-means implementation in faiss~\citep{faiss} which only takes $\sim10$ seconds.
The clustering is performed every epoch,
which introduces $1/3$ computational overhead due to a forward pass through the dataset.
The number of negatives for ProtoNCE is set as $k=16000$.
We also perform additional experiments by using a non-linear projection head (a 2-layer MLP) instead of a linear one to project the representations into a 128-D space, which has been shown to be beneficial for contrastive representation learning~\citep{simple}.
\fi

%% file: table/fig_framework.tex
\begin{figure*}[!t]
 \centering
   \includegraphics[width=\linewidth]{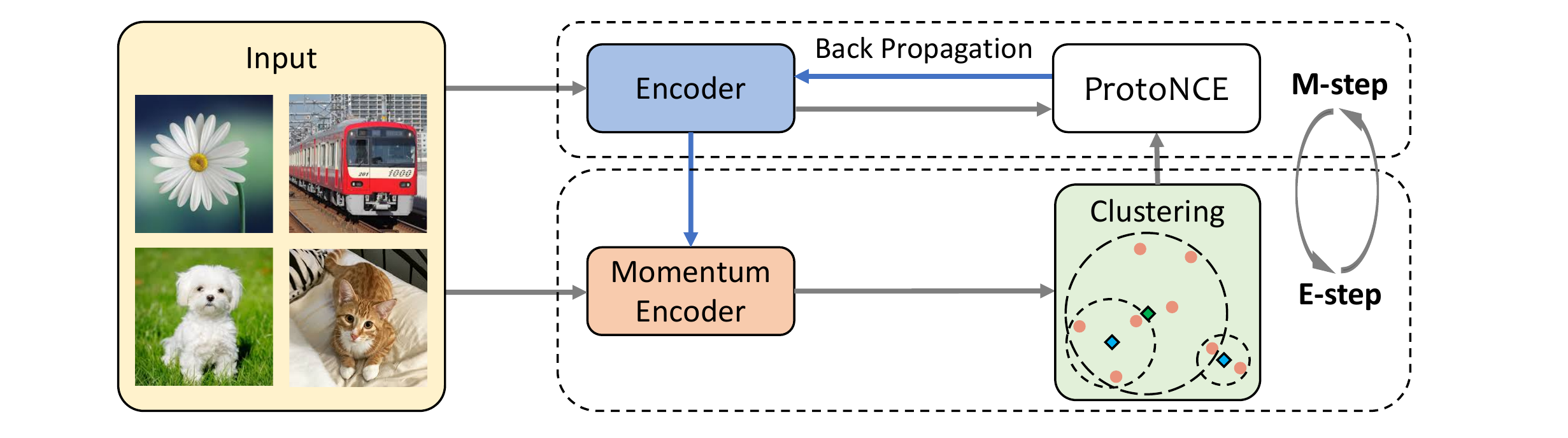}
   \vspace{-0.12in}
   \caption
  	{
  	\small
		Training framework of Prototypical Contrastive Learning. 
	} 
  \label{fig:framework}
     \vspace{-0.1in}
 \end{figure*}

%% file: table/fig_cluster_MI.tex
\begin{figure*}[!t]
 \centering
\small
 \begin{minipage}{0.49\textwidth}
	\centering
	\includegraphics[width=\textwidth]{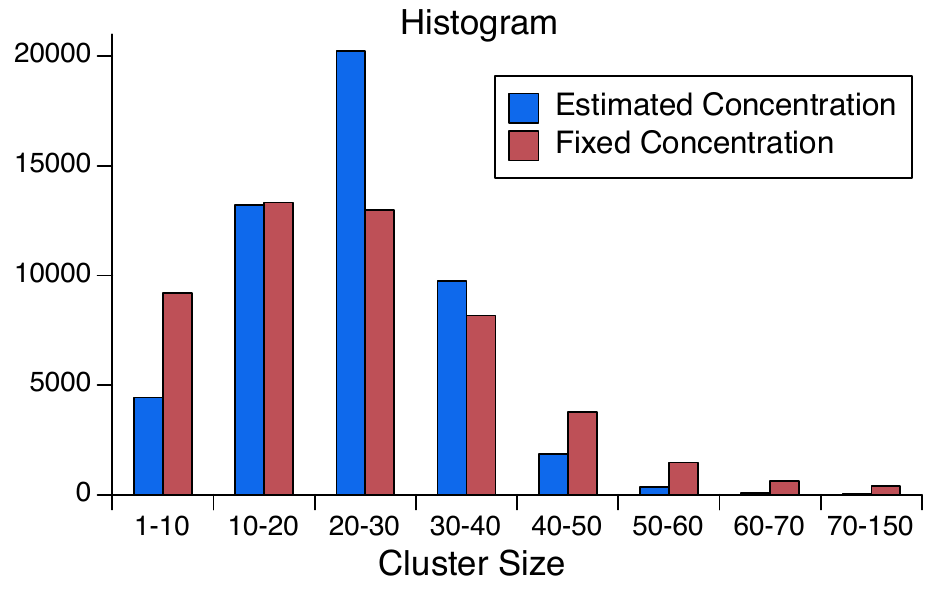}
	\small (a)
\end{minipage}
 \begin{minipage}{0.49\textwidth}
	\centering
	\includegraphics[width=\textwidth]{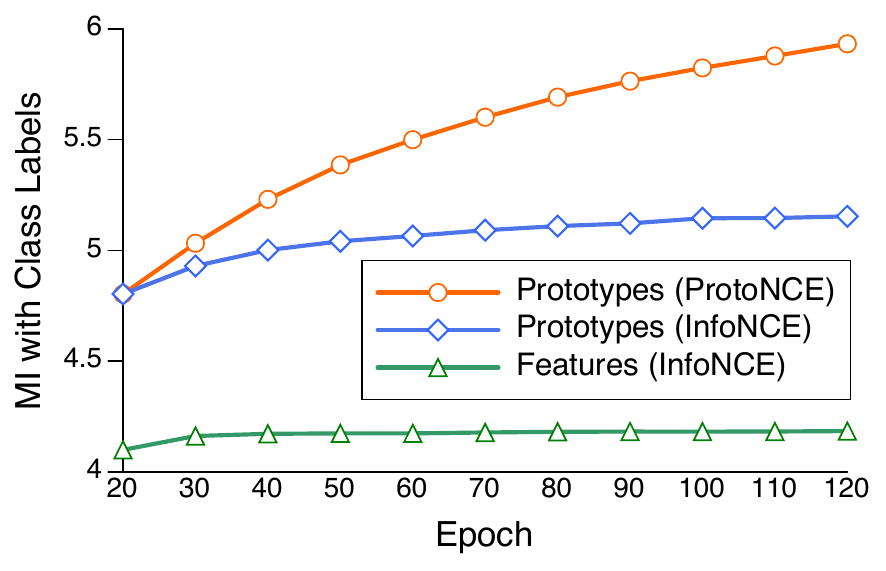}	
	\small (b) 
\end{minipage}
\vspace{-0.05in}
  \caption
  	{
  	\small
		(a) Histogram of cluster size for PCL ($\#$clusters $k$=50000) with fixed or estimated concentration.
		Using a different $\phi$ for each prototype yields more balanced clusters with similar sizes, which leads to better representation learning.
		(b) Mutual info between instance features (or their assigned prototypes) and class labels of all images in ImageNet. Compared to InfoNCE, our ProtoNCE learns better prototypes with more semantics. 
	  } 
 \vspace{-0.1in}
  \label{fig:cluster_MI}
 \end{figure*}

%% file: sec_experiment.tex
\vspace{-0.05in}
\section{Experiments}
\label{sec:experiment}
We evaluate PCL on transfer learning tasks,
based on the principle that good representations should transfer with limited supervision and fine-tuning. We follow the settings in MoCo,
therefore direct comparisons with MoCo could demonstrate the improvement from the prototypical contrastive loss.
%our method substantially outperforms MoCo across all tasks. 
%More details of our experimental setups and hyperparameters settings are given in appendices.
%More details of our method (pseudo-code, experimental settings, cluster analysis, and convergence proof) are given in appendix. Our code and models will be released.

\vspace{-0.05in}
\subsection{Implementation details}
\vspace{-0.05in}
% It has been shown that the performance of unsupervised learned representations can be improved by using a larger network, adopting stronger data augmentation, training for more epochs, or using a larger batchsize~\citep{mocov2,simple,CPCv2,PIRL}.
% However, these improvements usually come at the cost of more computation resources.
To enable a fair comparison,
we follow the same setting as MoCo.
%a recent state-of-the-art method for unsupervised representation learning.
We perform training on the ImageNet-1M dataset.
%which contains $\sim$1.28 million images in 1000 classes.
A ResNet-50~\citep{resnet} is adopted as the encoder,
whose last fully-connected layer outputs a 128-D and L2-normalized feature.
We follow previous works~\citep{moco,Instance} and perform data augmentation with random crop, random color jittering, 
random horizontal flip, and random grayscale conversion. 
We use SGD as our optimizer, with a weight decay of 0.0001, a momentum of 0.9, and a batch size of 256.
We train for 200 epochs, where we warm-up the network in the first 20 epochs by only using the InfoNCE loss.
The initial learning rate is 0.03, and is multiplied by 0.1 at 120 and 160 epochs.
In terms of the hyper-parameters,
we set $\tau=0.1$,
$\alpha=10$, $r=16000$, and number of clusters $K=\{25000,50000,100000\}$.
We also experiment with PCL v2 using improvements introduced by~\cite{simple,mocov2},
which includes a MLP projection layer, stronger data augmentation with additional Gaussian blur, and temperature $\tau=0.2$.
We adopt faiss~\citep{faiss} for efficient $k$-means clustering.
The clustering is performed per-epoch on center-cropped images.
We find over-clustering to be beneficial.
%More details of training settings can be found in Appendix A. 
Recent advances in self-supervised learning have been propelled by huge compute which is inaccessible to many researchers.
We instead target a more commonly accessible training resource for PCL with 4 NVIDIA-V100 GPUs and approximately 5 days of training.

\vspace{-0.05in}
\subsection{Image classification with limited training data}

%\subsection{Low-shot Classification}
{\bf Low-shot classification.} We evaluate the learned representation on image classification tasks with few training samples per-category. We follow the setup in~\cite{benchmark} and train linear SVMs using fixed representations on two datasets: Places205~\citep{places} for scene recognition and PASCAL VOC2007~\citep{voc} for object classification.
We vary the number $k$ of samples per-class and report the average result across 5 independent runs (standard deviation is reported in appendix~\ref{sec:std}). Table~\ref{tbl:lowshot_voc} shows the results, in which our method substantially outperforms both MoCo and SimCLR.

\begin{table*}[t!]
\small
	\centering	
	\setlength\tabcolsep{4.5pt}
	\resizebox{\textwidth}{!}{%
	\begin{tabular}	{l l |  r r r r r  | r r r r r}
		\toprule	 	
		%Method  & $k$=1 & $k$=2 & $k$=4 & $k$=8 &$k$=16 & $k$=1 & $k$=2 & $k$=4 & $k$=8 &$k$=16\\
		\multirow{2}{*}{Method}	& \multirow{2}{*}{architecture}  & \multicolumn{5}{c}{\bf VOC07} & \multicolumn{5}{c}{\bf Places205}\\
		& & $k$=1 & $k$=2 & $k$=4 & $k$=8 &$k$=16 & $k$=1 & $k$=2 & $k$=4 & $k$=8 &$k$=16\\
	    \midrule
	    Random  & \multirow{2}{*}{ResNet-50}&  8.0	&8.2	&8.2	&8.2	&8.5 & 0.7 &	0.7&0.7&0.7&0.7\\
	    Supervised & &54.3  & 67.8 &73.9 &79.6 & 82.3 &  14.9 & 21.0 &26.9 & 32.1 & 36.0 \\ \hline
	    
	    Jigsaw &\multirow{3}{*}{ResNet-50}& 26.5	&31.1	&40.0	&46.7	&51.8 & 4.6&	6.4&9.4&12.9	&17.4 \\
	    MoCo & & 31.4 &42.0 &49.5&60.0 &65.9 & 8.8 &13.2&18.2&23.2&28.0\\ 
	    PCL (ours)  & &\textbf{46.9} &\textbf{56.4}&\textbf{62.8}&\textbf{70.2}&\textbf{74.3} & \textbf{11.3} &\textbf{15.7}&\textbf{19.5}&\textbf{24.1}&\textbf{28.4} \\  
	    %{1: 11.338536585365855, 2: 15.744390243902439, 4: 19.45658536585366, 8: 23.773658536585366, 16: 27.807804878048778}
	    \hline
	    SimCLR&\multirow{3}{*}{ResNet-50-MLP} &32.7&43.1&52.5&61.0&67.1&9.4&14.2&19.3&23.7&28.3\\
	    MoCo v2& & 46.3 & 58.3 & 64.9 & 72.5 & 76.1 & 10.9 & 16.3 & 20.8 & 26.0 & 30.1\\   
	    PCL v2 (ours) & & \textbf{47.9} &\textbf{59.6}&\textbf{66.2}&\textbf{74.5}&\textbf{78.3} &\textbf{12.5} &\textbf{17.5}&\textbf{23.2}&\textbf{28.1}&\textbf{32.3} \\  	
		\bottomrule
	\end{tabular}
	}
		\vspace{-0.5ex}
	\caption
	{
		\small	
		\textbf{Low-shot image classification} on both VOC07 and Places205 datasets using linear SVMs trained on fixed representations. All methods were pretrained on ImageNet-1M dataset for 200 epochs (except for Jigsaw trained on ImageNet-14M). We vary the number of labeled examples $k$ and report the mAP (for VOC) and accuracy (for Places) across 5 runs. %Results for Jigsaw were taken from~\cite{benchmark}. 
		We use the released pretrained model for MoCo, and re-implement SimCLR.
	}
	\label{tbl:lowshot_voc}	
	\vspace{-1ex}
\end{table*}	

%\input{table/tbl_low_shot_voc}
%\input{table/tbl_low_shot_places}

%\subsection{Semi-supervised Image Classification}

{\bf Semi-supervised image classification.} 
We perform semi-supervised learning experiments to evaluate whether the learned representation can provide a good basis for fine-tuning.
Following the setup from~\cite{Instance,PIRL},
we randomly select a subset (1\% or 10\%) of ImageNet training data (with labels),
and fine-tune the self-supervised trained model on these subsets.
Table~\ref{tbl:semi_supervise} reports the top-5 accuracy on ImageNet validation set. 
Our method sets a new state-of-the-art under 200 training epochs, 
outperforming both self-supervised learning methods and semi-supervised learning methods.
The standard deviation across 5 runs is low ($<0.6$ for 1\% labels).

\input{table/tbl_semi_supervise}
\vspace{-0.5ex}
\subsection{Image classification benchmarks}
\vspace{-0.5ex}

{\bf Linear classifiers.} Next, we train linear classifiers on fixed image representations using the entire labeled training data.
We evaluate the performance of such linear classifiers on three datasets: ImageNet, VOC07, and Places205. 
Table~\ref{tbl:linear} reports the results.
PCL outperforms MoCo under direct comparison, which demonstrate the advantage of the proposed prototypical contrastive loss.

\input{table/tbl_linear_classify}

%\subsection{KNN Classification}
{\bf KNN classifiers.}
We perform k-nearest neighbor (kNN) classification on ImageNet.
For a query image with feature $v$, we take its top $k$ nearest neighbors from the momentum features,
and perform weighted-combination of their labels where the weights are calculated by $\mathrm{exp}(v\cdot v'_i / \tau)$. 
Table~\ref{tbl:knn} reports the accuracy. 
Our method substantially outperforms previous methods.

\input{table/tbl_linear_knn}

%\vspace{-4ex}
\subsection{Clustering evaluation}
\vspace{-1ex}
In Table~\ref{tbl:ami}, we evaluate the $k$-means clustering performance on ImageNet using representations learned by different methods.
PCL leads to substantially higher adjusted mutual information (AMI) score. Details are given in appendix~\ref{sec:ami}.

\vspace{-1ex}
\input{table/tbl_AMI}

\subsection{Object detection}
\vspace{-1ex}
We assess the representation on object detection. Following~\cite{benchmark},
we train a Faster R-CNN~\citep{faster} on VOC07 or VOC07+12,
and evaluate on the test set of VOC07.
%It has been shown that fine-tuning on randomly initialized models can achieve competitive results~\citep{rethink}.
We keep the pretrained backbone frozen to better evaluate the learned representation,
and use the same schedule for all methods.
Table~\ref{tbl:detection} reports the average mAP across three runs.
Our method substantially closes the gap between self-supervised methods and supervised training. 
In appendix~\ref{sec:coco},
we show the results for fine-tuning the pretrained model for object detection and instance segmentation on COCO~\citep{mscoco},
where PCL outperforms both MoCo and supervised training.

\input{table/tbl_detection}

\section{Visualization of learned representation}
In Figure~\ref{fig:tsne}, we visualize the unsupervised learned representation of ImageNet training images using t-SNE~\citep{tsne}.
Compared to the representation learned by MoCo,
the representation learned by the proposed PCL forms more separated clusters,
which also suggests representation of lower entropy.

\input{table/fig_tsne}

%% file: table/tbl_semi_supervise.tex
\begin{table*}[htb]
\small
	\centering	
	\setlength\tabcolsep{5pt}
	%\resizebox{\textwidth}{!}{%
	\begin{tabular}	{l   l  l  |c |c  }
		\toprule	 	
		\multirow{2}{*}{Method}	  & \multirow{2}{*}{architecture}& \#pretrain& \multicolumn{2}{c}{Top-5 Accuracy}\\
		& & epochs &1\% &  10\%\\
		\midrule
		Random~\citep{Instance} & ResNet-50 & - &22.0  &59.0\\
		Supervised baseline~\citep{S4L}  & ResNet-50 & -& 48.4 & 80.4\\
		\hline
		\multicolumn{2}{l}{\textit{Semi-supervised learning methods:}} & & &\\
		\hline
		Pseudolabels~\citep{S4L}  & ResNet-50v2& - &	51.6 &  82.4\\
		VAT + Entropy Min.~\citep{VAT}  &ResNet-50v2 &  -& 	47.0 & 83.4\\
		%S$^4$L Exemplar~\citep{S4L}  &ResNet-50v2&  - & 	47.0 & 83.7\\
		S$^4$L Rotation~\citep{S4L}   &ResNet-50v2&  -& 	53.4& 83.8\\	
		\hline	
		\multicolumn{2}{l}{\textit{Self-supervised learning methods:}} & & &\\
	    \hline			
		Instance Discrimination~\citep{Instance}  & ResNet-50& 200 &	39.2& 77.4\\
		Jigsaw~\citep{Noroozi_ECCV_2016} &ResNet-50& 90  & 45.3 & 79.3\\			SimCLR~\citep{simple}  &ResNet-50-MLP &  200 & 56.5& 82.7\\	 
		MoCo~\citep{moco}   &ResNet-50&  200& 56.9   & 83.0 \\		 
			
		MoCo v2~\citep{mocov2}   &ResNet-50-MLP&  200& 66.3   & 84.4 \\
		PCL v2 (ours) & ResNet-50-MLP&  200 & 73.9  & 85.0\\
		PCL (ours) & ResNet-50&  200 & \textbf{75.3}  & \textbf{85.6}\\
		\hline
		PIRL~\citep{PIRL}   &ResNet-50&  800& 57.2 &83.8 \\
		SimCLR~\cite{simple}& ResNet-50-MLP& 1000 & ~75.5$^\dagger$ &~87.8$^\dagger$ \\
		BYOL~\citep{byol} & ResNet-50-MLP$_\mathrm{big}$&  1000& ~78.4$^\dagger$ &~89.0$^\dagger$ \\
		SwAV~\citep{swav} &ResNet-50-MLP & 800& ~78.5$^\ddagger$ & ~89.9$^\ddagger$\\
		
  %\hline		  
%SimCLR~\citep{simple}  &ResNet-50-MLP &  1000& ~75.5$^\dagger$ & ~87.8$^\dagger$ \\	  
		%PCL-MLP  &ResNet-50-MLP&  200& \textbf{}   & \\ 	 		
		\bottomrule
	\end{tabular}
	%}
	\vspace{-0.5ex}
	\caption
	{
		\small	
		\textbf{Semi-supervised learning} on ImageNet. We report top-5 accuracy on the ImageNet validation set of self-supervised models that are
		finetuned on 1\% or 10\% of labeled data. %We use the released pretrained model for MoCo, and re-implement SimCLR; all other numbers are adopted from corresponding papers.
		%SimCLR$^\dagger$ requires a high computation budget, with a large batch size of 4096 trained on 128 TPUs for 1000 epochs.
		$^\ddagger$: SimCLR, BYOL, and SwAV use a large batch size of 4096.		
		$^\ddagger$: SwAV uses multi-crop augmentation.
		}
	\label{tbl:semi_supervise}	
	\vspace{-1ex}
\end{table*}		

%% file: table/tbl_linear_classify.tex
\begin{table*}[hptb]
\small
	\vspace{-0.05in}
	\centering	
	\setlength\tabcolsep{4pt}
	\resizebox{1\textwidth}{!}{%
	\begin{tabular}	{l  |  l  | l | |c |c | c }
		\toprule	 	
    	\multirow{2}{*}{Method}	 & architecture&\#pretrain  & \multicolumn{3}{c}{Dataset}\\
		& (\#params)&epochs& ImageNet& VOC07 & Places205 \\
		\midrule
		%Colorization~\citep{Zhang_ECCV_2016} & R50 (24M) & 28 & 39.6 & 55.6 & 37.5\\
		Jigsaw~\citep{Noroozi_ECCV_2016} & R50 (24M) & 90  & 45.7 & 64.5 & 41.2\\
		Rotation~\citep{Gidaris_2018_ICLR}& R50 (24M) &– &48.9 &63.9& 41.4 \\		
		DeepCluster~\citep{deepcluster} & VGG(15M) &100& 48.4 & 71.9& 37.9\\
		BigBiGAN~\citep{bigan} &  R50 (24M) &  –& 56.6 & –& –\\
		InstDisc~\citep{Instance}& R50 (24M) & 200  & 54.0 & –& 45.5\\
		MoCo~\citep{moco}& R50 (24M) & 200  & 60.6 & ~79.2$^*$& ~48.9$^*$\\
		PCL (ours) & R50 (24M) & 200  & \textbf{61.5}&\textbf{82.3} &\textbf{49.2}\\
		\hline
		SimCLR~\citep{simple} & R50-MLP (28M) & 200 & 61.9 & – & – \\
		MoCo v2~\citep{mocov2} & R50-MLP (28M) & 200 & 67.5 & ~84.0$^*$ & ~50.1$^*$\\
		PCL v2 (ours) & R50-MLP (28M) & 200  & \textbf{67.6}& \textbf{85.4} & \textbf{50.3}\\
		%83.7
	    \hline 
		LocalAgg~\citep{local}& R50 (24M) & 200  & ~60.2$^\dagger$ &  – & ~50.1$^\dagger$\\	 
		SelfLabel~\citep{Sela} & R50 (24M) & 400  & 61.5 &  –&  – \\	   
	    CPC~\citep{CPC} & R101 (28M) &  – &48.7 &  –&  –\\
	    %CPCv2~\citep{CPCv2} & R170$_{w}$ (303M) & $\sim$200 &65.9 & – &  –\\
	    CMC~\citep{CMC} & R50$_{L+ab}$ (47M) & 280 & 64.0 &  – &  – \\
	    PIRL~\citep{PIRL} & R50 (24M) & 800  &63.6 &81.1 &49.8\\
	    AMDIM~\citep{AMDIM} & Custom (626M) & 150& ~68.1$^\dagger$ & – & ~55.0$^\dagger$ \\	    
	    SimCLR~\citep{simple} & R50-MLP (28M) & 1000  & ~69.3$^\dagger$ & ~80.5$^\dagger$ &  – \\
	    %SimCLR~\citep{simple} & R50-$4\times$(375M) & 1000  & ~76.5$^\dagger$ & ~84.2$^\dagger$ & - \\
	    BYOL~\citep{byol} & R50-MLP$_\mathrm{big}$(35M) & 1000  & ~74.3$^\dagger$ & - &  – \\
	    SwAV~\citep{swav} & R50-MLP (28M) & 800  & ~75.3$^\dagger$ & ~88.9$^\dagger$ &  ~56.7$^\dagger$ \\
		\bottomrule
	\end{tabular}
	}
	\vspace{-0.5ex}
	\caption
	{
		\small	
		\textbf{Image classification with linear models.} We report top-1 accuracy. Numbers with $^*$ are from released pretrained model; all other numbers are adopted from corresponding papers.\\
		{$^\dagger$: LocalAgg uses 10-crop evaluation. ADMIM uses FastAutoAugment~\citep{fastaug} that is supervised by ImageNet labels.
		SwAV uses multi-crop augmentation.
		SimCLR, BYOL, and SwAV use a large batch size of 4096.}
	}
	\label{tbl:linear}	
	%\vspace{-1ex}
\end{table*}		

%% file: table/tbl_linear_knn.tex
\begin{table*}[!hb]
%	\vspace{-0.05in}
	\centering	
	\setlength\tabcolsep{5pt}
	\resizebox{1\textwidth}{!}{%
	\begin{tabular}	{l  | c | c|c |c }
		\toprule	 	
		Method   & Inst. Disc.~\citep{Instance}&MoCo~\citep{moco} &  LA~\citep{local} &  PCL (ours)\\
		\midrule
	Accuracy &  46.5& 47.1& 49.4 & \textbf{54.5}\\
		\bottomrule
	\end{tabular}
	}
		\vspace{-0.1in}
	\caption
	{
		\small	
		\textbf{Image classification with kNN classifiers} using ResNet-50 features on ImageNet. %We report top-1 accuracy. Results for Inst. Disc and LA are taken from corresponding papers. 
		%Result for MoCo is from released model.
		%Our method only requires 20 nearest neighbors, as opposed to 200 in~\citep{Instance,local}. 
	}
	\label{tbl:knn}	
	%\vspace{-0.05in}
\end{table*}

%% file: table/tbl_AMI.tex
\begin{table*}[!ht]
\small
%	\vspace{-0.05in}
	\centering	
	\setlength\tabcolsep{5pt}
	%\resizebox{0.75\textwidth}{!}{%
	\begin{tabular}	{l  | c | c|c }
		\toprule	 	
		Method   & DeepCluster~\citep{deepcluster}&MoCo~\citep{moco} &  PCL (ours)\\
		\midrule
	AMI & 0.281& 0.285& \textbf{0.410}\\
		\bottomrule
	\end{tabular}
	%}
		\vspace{-0.05in}
	\caption
	{
		\small	
		AMI score for k-means clustering ($k=25000$) on ImageNet representation.
	}
	\vspace{-0.1in}
	\label{tbl:ami}	
\end{table*}		

%% file: table/tbl_detection.tex
\begin{table*}[htpb]
\small
	\centering	
	\setlength\tabcolsep{5pt}
	%\resizebox{\textwidth}{!}{%
	\begin{tabular}	{l | l |  l | c | c}
		\toprule	 	
		Method& Pretrain Dataset   & Architecture &  \multicolumn{2}{c}{Training data}\\
		& & & ~VOC07~ & VOC07+12\\
			
	    \midrule
	    %Supervised~\citep{benchmark} &  ImageNet-1M & Resnet-50-C4 &67.1 & 68.3 \\
	    Supervised & ImageNet-1M  & Resnet-50-FPN &72.8 & 79.3 \\
	    \hline
	    %Jigsaw~\citep{Noroozi_ECCV_2016} &  ImageNet-14M & Resnet-50-C4  &62.7 & 64.8 \\
	    MoCo~\citep{moco} &  ImageNet-1M & Resnet-50-FPN &66.4&  73.5\\
	    PCL (ours) &  ImageNet-1M& Resnet-50-FPN  & \textbf{71.7} &\textbf{78.5}\\
		\bottomrule
	\end{tabular}
	%}
	\vspace{-0.05in}
	\caption
	{
		\small	
		\textbf{Object detection} for frozen \textsf{conv} body on VOC using Faster R-CNN.
		%We measure the average mAP$@0.5$ on VOC07 test set across three runs.
	}
	\label{tbl:detection}
		\vspace{-0.1in}
\end{table*}

%% file: table/fig_tsne.tex
\begin{figure*}[!ht]
   \centering
    \begin{minipage}{0.49\textwidth}
	    \centering
	    \includegraphics[width=\textwidth,trim={0.4cm 0 0 0},clip]{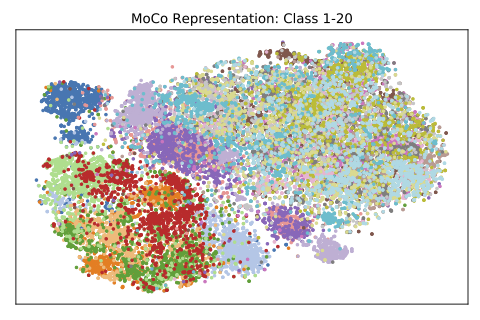}
    \end{minipage}
     \begin{minipage}{0.49\textwidth}
    	\centering
    	\includegraphics[width=\textwidth,trim={0 0 0.3cm 0},clip]{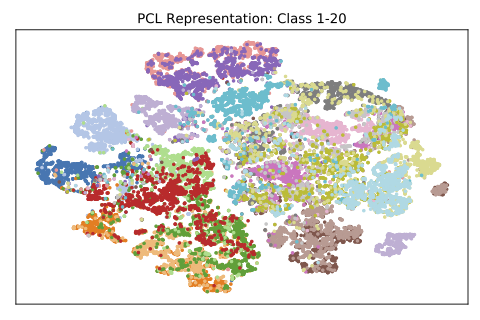}	
    \end{minipage}
    \begin{minipage}{0.49\textwidth}
	    \centering
	    \includegraphics[width=\textwidth,trim={0.4cm 0 0 0},clip]{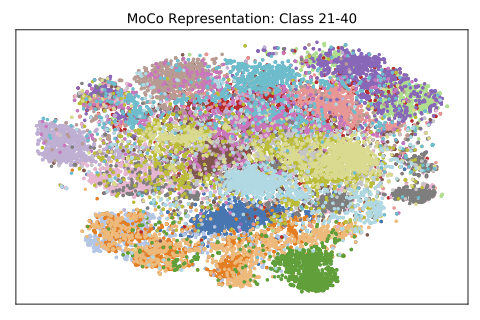}
    \end{minipage}
     \begin{minipage}{0.49\textwidth}
    	\centering
    	\includegraphics[width=\textwidth,trim={0 0 0.3cm 0},clip]{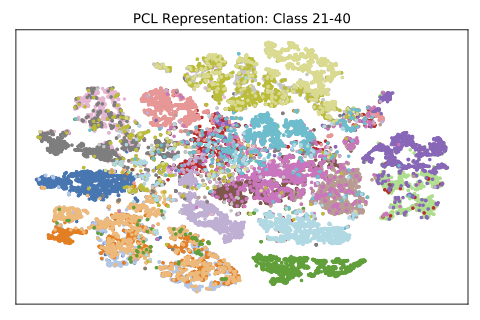}	
    \end{minipage}
    % \begin{minipage}{0.49\textwidth}
	   % \centering
	   % \includegraphics[width=\textwidth,trim={0.4cm 0 0 0},clip]{moco_tsne_41-60.png}
    % \end{minipage}
    %  \begin{minipage}{0.49\textwidth}
    % 	\centering
    % 	\includegraphics[width=\textwidth,trim={0 0 0.3cm 0},clip]{pcl_tsne_41-60.png}	
   % \end{minipage}    
   \caption
  	{
  	\small
		T-SNE visualization of the unsupervised learned representation for ImageNet training images from the first 40 classes. Left: MoCo; Right: PCL (ours). Colors represent classes.
	} 
  \label{fig:tsne}
 \end{figure*}

%% file: sec_conclusion.tex
\vspace{-0.05in}
\section{Conclusion}
\vspace{-0.05in}
\label{sec:conclusion}
This paper proposes Prototypical Contrastive Learning,
a generic unsupervised representation learning framework that finds network parameters to maximize the log-likelihood of the observed data. We introduce prototypes as latent variables, and perform iterative clustering and representation learning in an EM-based framework. PCL learns an embedding space which encodes the semantic structure of data,
by training on the proposed ProtoNCE loss.
Our extensive experiments on multiple benchmarks demonstrate the advantage of PCL for unsupervised representation learning.

% Advances in unsupervised representation learning will impact our society in many ways.
% PCL enables training deep models that can be transferred to many visual perception tasks,
% which leads to smarter AI systems with better capability to help human.
% Since unsupervised learning does not require the expensive process of manual annotation,
% AI practitioners could benefit by training models at much lower costs.
% However, there is the concern that crowd-workers may have reduced income due to less need for their annotation service.
% Another ethical issue is the potential risk that people's private photos posted online my be used as training data.
% Enhanced privacy protection measures needed to be taken to eliminate such risk.

%% file: sec_appendix.tex
\newpage

\begin{appendices}

%\section{Training details for unsupervised learning}
% For the unsupervised learning experiment,
% We follow previous works~\citep{moco,Instance} and perform data augmentation with random crop, random color jittering, 
% random horizontal flip, and random grayscale conversion. 
% We use SGD as our optimizer, with a weight decay of 0.0001, a momentum of 0.9, and a batch size of 256.
% We train for 200 epochs, where we warm-up the network in the first 20 epochs by only using the InfoNCE loss.
% The initial learning rate is 0.03, and is multiplied by 0.1 at 120 and 160 epochs.
% In terms of the hyper-parameters,
% we set $\tau=0.1$,
% $\alpha=10$, $r=16000$, and number of clusters $K=\{25000,50000,100000\}$.
% For PCL v2, we follow~\cite{simple,mocov2} and use a MLP projection layer, stronger data augmentation with additional Gaussian blur, and temperature $\tau=0.2$.
% The clustering is performed per-epoch on center-cropped images.
% We find over-clustering to be beneficial.
% We use the GPU $k$-means implementation in faiss~\citep{faiss} which takes less than 20 seconds.
% Overall, PCL introduces $\sim 1/3$ computational overhead compared to MoCo.
% \vspace{-0.05in}
\section{Ablation on ProtoNCE}
\vspace{-0.05in}
The proposed loss in eqn.(\ref{eqn:proto_NCE}) contains two terms:
the instance-wise contrastive loss and the proposed prototypical contrastive loss.
Here we study the effect of each term on representation learning.
Table~\ref{tbl:ablation} reports the results for low-resource fine-tuning and linear classification on ImageNet.
The prototypical term plays an important role, especially in the fine-tuning experiment.
The warm-up also improves the result by bootstrapping the clustering with better representations.

\begin{table*}[!ht]
	\centering	
	\setlength\tabcolsep{5pt}
	%\resizebox{0.75\textwidth}{!}{%
	\begin{tabular}	{l  | c | c }
		\toprule	 	
		Method   & 1\% fine-tuning (top-5 acc.) &  linear classification (top-1 acc.)\\
		\midrule
    	instance only & 56.9 & 60.6 \\
    	proto only (w/o warm-up) & 60.7 & 60.4 \\
    	proto only (w/ warm-up) & 72.3 & 60.9\\
    	instance + proto (w/o warm-up) & 74.6 & 61.3 \\
    	instance + proto (w/ warm-up) & \textbf{75.3} & \textbf{61.5}\\
		\bottomrule
	\end{tabular}
	%}
	\caption
	{
		\small	
		Effect of instance-wise contrastive loss and prototypical contrastive loss.
	}
	\label{tbl:ablation}	
\end{table*}		
\vspace{-0.05in}

\section{Pseudo-code for Prototypical Contrastive Learning}
\label{sec:pseudo_code}
\vspace{-0.05in}
\input{table/Alg}

\section{Standard deviation for low-shot classification}
\label{sec:std}

In Table~\ref{tbl:ablation},
we report the standard deviation for the low-shot classification experiments in Table~\ref{tbl:lowshot_voc}.

\begin{table*}[ht!]
%\small
	\centering	
	\setlength\tabcolsep{4.5pt}
	\begin{tabular}	{l |  r r r r r  | r r r r r}
		\toprule	 	
		%Method  & $k$=1 & $k$=2 & $k$=4 & $k$=8 &$k$=16 & $k$=1 & $k$=2 & $k$=4 & $k$=8 &$k$=16\\
		\multirow{2}{*}{Method}	&  \multicolumn{5}{c}{\bf VOC07} & \multicolumn{5}{c}{\bf Places205}\\
		&  $k$=1 & $k$=2 & $k$=4 & $k$=8 &$k$=16 & $k$=1 & $k$=2 & $k$=4 & $k$=8 &$k$=16\\
	    \midrule
	    PCL&  4.06 & 2.65 & 2.21 & 0.49 & 0.39 & 0.24 & 0.23 & 0.13 & 0.07 & 0.05\\   
	    PCL v2&  4.12 & 2.70 & 2.17 & 0.54 & 0.38 & 0.26 & 0.23 & 0.12 & 0.08 & 0.04\\   
    \bottomrule
	\end{tabular}
	\caption
	{
		\small	
	    Standard deviation across 5 runs for low-shot image classification experiments. 
	}
	\label{tbl:lowshot_std}	
\end{table*}	

\section{COCO object detection and segmentation}
\label{sec:coco}
Following the experiment setting in~\citep{moco}, we use Mask R-CNN~\citep{mask_rcnn} with C4 backbone. 
We finetune all layers end-to-end on the COCO train2017 set and evaluate
on val2017. The schedule is the default $2\times$ in~\citep{Detectron2018}.
PCL outperforms both MoCo~\citep{moco} and supervised pre-training in all metrics.
\input{table/tbl_coco}

\section{Training details for transfer learning experiments}

For training linear SVMs on Places and VOC,
we follow the procedure in~\citep{benchmark} and use the LIBLINEAR~\citep{liblinear} package.
We preprocess all images by resizing to 256 pixels along the shorter side and taking a $224\times224$ center crop.
The linear SVMs are trained on the global average pooling features of ResNet-50.

For image classification with linear models,
we use the pretrained representations from the global average pooling features (2048-D) for ImageNet and VOC, and the conv5 features (averaged pooled to $\sim$9000-D) for Places.
We train a linear SVM for VOC, and a logistic regression classifier (a fully-connected layer followed by softmax) for ImageNet and Places.
The logistic regression classifier is trained using SGD with a momentum of 0.9.
For ImageNet, we train for 100 epochs with an initial learning rate of 10 and a weight decay of 0. 
Similar hyper-parameters are used by~\citep{moco}.
For Places, we train for 40 epochs with an initial learning rate of 0.3 and a weight decay of 0.

For semi-supervised learning,
we finetune ResNet-50 with pretrained weights on a subset of ImageNet with labels.
We optimize the model with SGD, using a batch size of 256, a momentum of 0.9, and a weight decay of 0.0005.
We apply different learning rate to the ConvNet and the linear classifier.
The learning rate for the ConvNet is 0.01, and the learning rate for the classifier is 0.1 (for 10\% labels) or 1 (for 1\% labels).
We train for 20 epochs,
and drop the learning rate by 0.2 at 12 and 16 epochs.

For object detection on VOC,
We use the R50-FPN backbone for the Faster R-CNN detector available in the \texttt{MMdetection}~\citep{mmdetection} codebase.
We freeze all the \textsf{conv} layers and also fix the BatchNorm parameters.
The model is optimized with SGD, using a batch size of 8, a momentum of 0.9, and a weight decay of 0.0001.
The initial learning rate is set as 0.05.
We finetune the models for 15 epochs,
and drop the learning rate by 0.1 at 12 epochs.

%\subsection{Ablation Study}
%\noindent\textbf{Effect of the instance discrimination loss.}
%\noindent\textbf{Effect of the number of negative samples.}

\section{Evaluation of Clustering}
\label{sec:ami}
In order to evaluate the quality of the clusters produced by PCL,
we compute the adjusted mutual information score (AMI)~\citep{AMI} between the clusterings and the ground-truth labels for ImageNet training data.
AMI is adjusted for chance which accounts for the bias in MI to give high values to clusterings with a larger number of clusters.
AMI has a value of 1 when two partitions are identical, and an expected value of 0 for random (independent) partitions.
In Figure~\ref{fig:AMI}, we show the AMI scores for three clusterings obtained by PCL, with number of clusters $K=\{25000,50000,100000\}$.
In Table~\ref{tbl:ami}, we show that compared to DeepCluster~\citep{deepcluster} and MoCo~\citep{moco},
PCL produces clusters of substantially higher quality.

\input{table/fig_AMI}

\section{Convergence proof}

Here we provide the proof that the proposed PCL would converge. Suppose let 
\begin{equation}
\begin{split}
F(\theta) = \sum_{i=1}^n \log p(x_i;\theta) 
=  \sum_{i=1}^n \log \sum_{c_i \in C}  p(x_i,c_i;\theta)  
& = \sum_{i=1}^n \log \sum_{c_i \in C} Q(c_i) \frac{p(x_i,c_i;\theta)}{Q(c_i)}\\
& \geq \sum_{i=1}^n \sum_{c_i \in C} Q(c_i)  \log \frac{p(x_i,c_i;\theta)}{Q(c_i)}.
\end{split}
\end{equation}
We have shown in Section~\ref{sec:em} that the above inequality holds with equality when $Q(c_i) = p(c_i ; x_i,\theta)$.

At the $t$-th E-step, we have estimated $Q^t(c_i) = p(c_i ; x_i,\theta^t)$. Therefore we have:
\begin{equation}
\label{eqn:f}
F(\theta^t) = \sum_{i=1}^n \sum_{c_i \in C} Q^t(c_i)  \log \frac{p(x_i,c_i;\theta^{t})}{Q^t(c_i)}.
\end{equation}

At the $t$-th M-step, we fix $Q^t(c_i) = p(c_i ; x_i,\theta^t)$ and train parameter $\theta$ to maximize Equation~\ref{eqn:f}.
Therefore we always have:
\begin{equation}
F(\theta^{t+1}) \geq \sum_{i=1}^n \sum_{c_i \in C} Q^{t}(c_i)  \log \frac{p(x_i,c_i;\theta^{t+1})}{Q^{t}(c_i)}
\geq \sum_{i=1}^n \sum_{c_i \in C} Q^t(c_i)  \log \frac{p(x_i,c_i;\theta^t)}{Q^t(c_i)} = F(\theta^{t}).
\end{equation}

The above result suggests that $F(\theta^{t})$ monotonously increase along with more iterations. Hence the algorithm will converge.

\section{Visualization of clusters}

In Figure~\ref{fig:example},
we show ImageNet training images that are randomly chosen from clusters generated by the proposed PCL.
PCL not only clusters images from the same class together,
but also finds fine-grained patterns that distinguish sub-classes,
demonstrating its capability to learn useful semantic representations.

\input{table/fig_example}

\end{appendices}

%% file: table/Alg.tex
\newcommand\mycommfont[1]{\footnotesize\ttfamily\textcolor{PineGreen}{#1}}
\SetCommentSty{mycommfont}
\begin{algorithm}[h]
	
	\DontPrintSemicolon
	\SetNoFillComment
	\small
	\textbf{Input:} encoder $f_\theta$, training dataset $X$, number of clusters $K=\{k_m\}_{m=1}^M$ \\	
    $\theta'=\theta$	\tcp*{initialize momentum encoder as the encoder}
	\While{$\mathrm{not~MaxEpoch}$}   
	{
	\tcc{E-step}
	$V'=f_{\theta'}(X)$ \tcp*{get momentum features for all training data}
	\For {$m=1$ \KwTo $M$}
	{
		$C^m=k\mathrm{-means}(V',k_m)$ \tcp*{cluster $V'$ into $k_m$ clusters, return prototypes}
		$\phi_m=\mathrm{Concentration}(C^m,V')$ \tcp*{estimate the distribution concentration around each prototype with Equation~\ref{eqn:mu}}
	} 
	\tcc{M-step}
	\For ( \tcp*[f]{load a minibatch $x$}) {$x~\mathbf{in}~\mathrm{Dataloader}(X)$}	
	{
		$v=f_\theta(x), v'=f_{\theta'}(x)$ \tcp*{forward pass through encoder and momentum encoder}
		%$v'=f_{\theta_m}(x)$ \tcp*{forward pass through momentum encoder}
		$\mathcal{L}_\mathrm{ProtoNCE}(v,v',\{C^m\}_{m=1}^M,\{\phi_m\}_{m=1}^M)$ \tcp*{calculate loss with Equation~\ref{eqn:proto_NCE}}
		$\theta=\mathrm{SGD}(\mathcal{L}_\mathrm{ProtoNCE},\theta)$  \tcp*{update encoder parameters}
		$\theta' = 0.999*\theta' + 0.001*\theta$ \tcp*{update momentum encoder}
	}    	
}
	\caption{\small Prototypical Contrastive Learning.}
\label{alg:pcl}
\end{algorithm}

%% file: table/tbl_coco.tex
\begin{table*}[!ht]

	\centering	
	\setlength\tabcolsep{5pt}
	%\resizebox{\textwidth}{!}{%
	\begin{tabular}	{l  |  c  c  c | c  c  c}
		\toprule	 	
		Method & AP$^\text{bb}$&   AP$^\text{bb}_{50}$ &  AP$^\text{bb}_{75}$ & AP$^\text{mk}$&   AP$^\text{mk}_{50}$ &  AP$^\text{mk}_{75}$\\
	    \midrule
	    Supervised & 40.0 & 59.9 & 43.1 & 34.7 & 56.5 &36.9 \\
	    MoCo~\citep{moco} & 40.7 & 60.5 & 44.1 & 35.4 & 57.3& 37.6 \\
	    PCL (ours) &  \textbf{41.0} &\textbf{60.8} & \textbf{44.2}&  \textbf{35.6} &\textbf{57.4} & \textbf{37.8}\\
		\bottomrule
	\end{tabular}
	%}
	\caption
	{
		\small	
		Object detection and instance segmentation fine-tuned on COCO.
		We evaluate bounding-box AP (AP$^\text{bb}$) and mask AP (AP$^\text{mk}$) on val2017.
	}
	\label{tbl:coco}

\end{table*}		

%% file: table/fig_AMI.tex
\begin{figure*}[!t]
 \centering
   \includegraphics[width=0.6\linewidth]{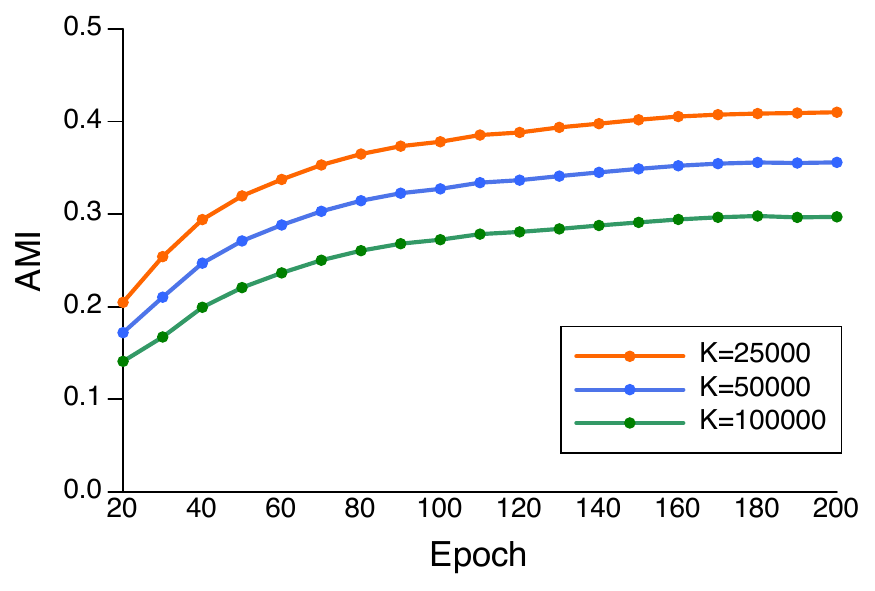}
   \caption
  	{
  	\small
		Adjusted mutual information score between the clusterings generated by PCL and the ground-truth labels for ImageNet training data.
	} 
  \label{fig:AMI}
 \end{figure*}

%% file: table/fig_example.tex
\begin{figure*}[h]
 \centering
   \includegraphics[width=\linewidth]{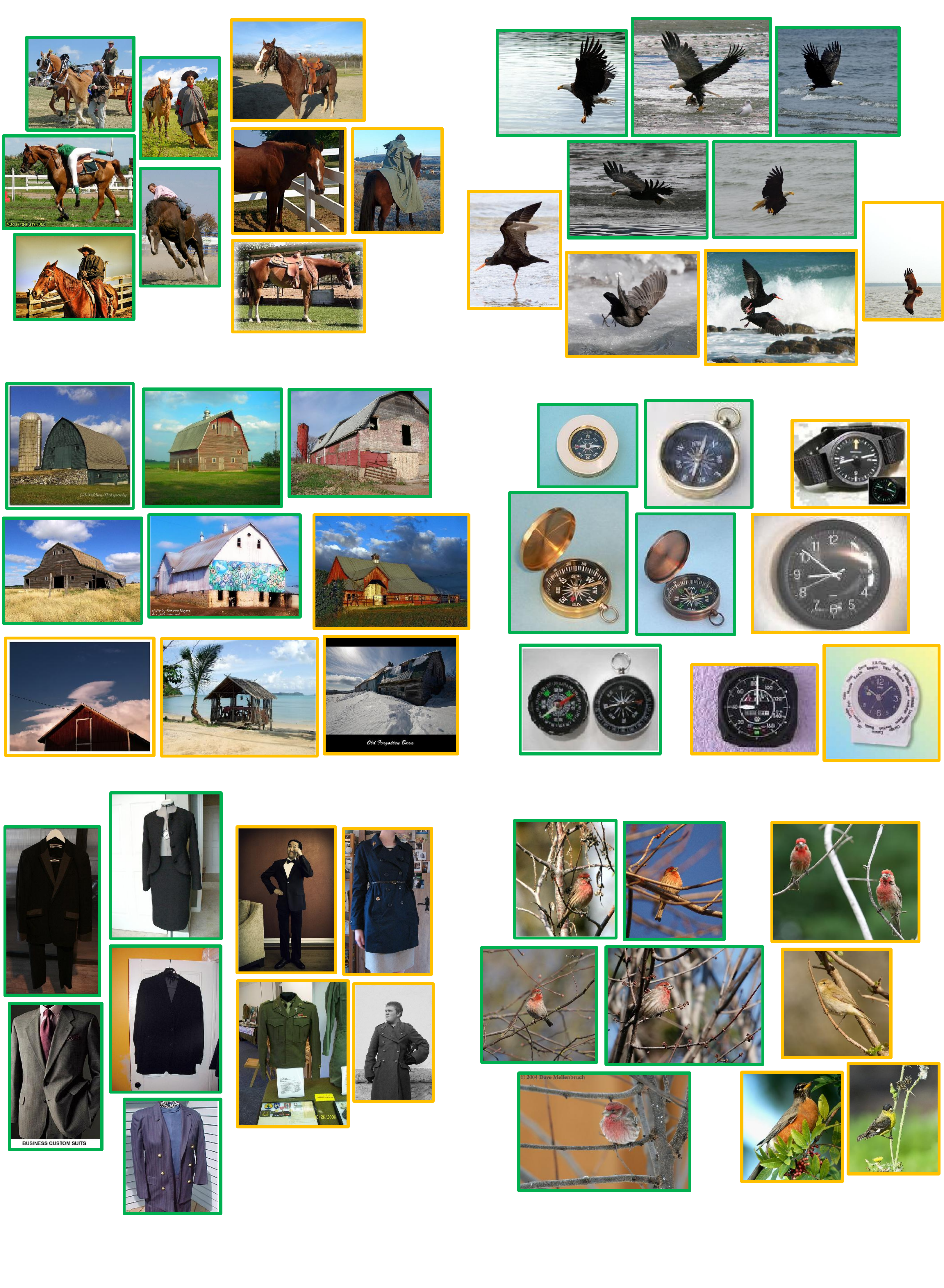}
   \vspace{-0.3in}
   \caption
  	{
  	\small
  	Visualization of randomly chosen clusters generated by PCL. {\color{ForestGreen}Green} boarder marks top-5 images that are closest to fine-grained prototypes ($K=100k$). \textcolor{orange}{Orange} boarder marks images randomly chosen from coarse-grained clusters ($K=50k$) that also cover the same green images.
  	PCL can discover hierarchical semantic structures within the data (\eg~images with horse and man form a fine-grained cluster within the coarse-grained horse cluster.)
	} 
  \label{fig:example}
 \end{figure*}